\documentclass[9.5pt,cspaper,journal,compsoc]{IEEEtran}
\usepackage{multirow}
\usepackage{booktabs}
\usepackage{float}
\usepackage{color}
\ifCLASSOPTIONcompsoc
  \usepackage[nocompress]{cite}
\else
  \usepackage{cite}
\fi
\ifCLASSINFOpdf
  \usepackage[pdftex]{graphicx}
\else
  \usepackage[dvips]{graphicx}
\fi
\usepackage{amsmath}
\usepackage{amssymb}
\usepackage{url}

\begin{document}
%
\title{Cluster-Level Contrastive Learning for Emotion Recognition in Conversations}
%
%
%
%

\author{Kailai~Yang,~\IEEEmembership{}
        Tianlin~Zhang,~\IEEEmembership{}
        Hassan~Alhuzali,~\IEEEmembership{}
        Sophia~Ananiadou~\IEEEmembership{}
\IEEEcompsocitemizethanks{\IEEEcompsocthanksitem Kailai Yang, Tianlin Zhang, and Sophia Ananiadou are with NaCTeM and the
Department of Computer Science, The University of Manchester,  United
Kingdom.\protect\\
E-mail: kailai.yang@postgrad.manchester.ac.uk\\
E-mail: tianlin.zhang@postgrad.manchester.ac.uk\\
E-mail: sophia.ananiadou@manchester.ac.uk
\IEEEcompsocthanksitem Hassan Alhuzali is with the College of Computers and Information Systems, Umm Al-Qura University, Saudi Arabia.\\
E-mail: hrhuzali@uqu.edu.sa}
}

%
%

\markboth{Journal of \LaTeX\ Class Files,~Vol.~14, No.~8, August~2015}%
{Shell \MakeLowercase{\textit{et al.}}: Bare Demo of IEEEtran.cls for Computer Society Journals}
%



\IEEEtitleabstractindextext{%
\begin{abstract}
A key challenge for Emotion Recognition in Conversations (ERC) is to distinguish semantically similar emotions. Some works utilise Supervised Contrastive Learning (SCL) which uses categorical emotion labels as supervision signals and contrasts in high-dimensional semantic space. However, categorical labels fail to provide quantitative information between emotions. ERC is also not equally dependent on all embedded features in the semantic space, which makes the high-dimensional SCL inefficient. To address these issues, we propose a novel low-dimensional Supervised Cluster-level Contrastive Learning (SCCL) method, which first reduces the high-dimensional SCL space to a three-dimensional affect representation space Valence-Arousal-Dominance (VAD), then performs cluster-level contrastive learning to incorporate measurable emotion prototypes. To help modelling the dialogue and enriching the context, we leverage the pre-trained knowledge adapters to infuse linguistic and factual knowledge. Experiments show that our method achieves new state-of-the-art results with $69.81\%$ on IEMOCAP, $65.7\%$ on MELD, and $62.51\%$ on DailyDialog datasets. The analysis also proves that the VAD space is not only suitable for ERC but also interpretable, with VAD prototypes enhancing its performance and stabilising the training of SCCL. In addition, the pre-trained knowledge adapters benefit the performance of the utterance encoder and SCCL. Our code is available at: https://github.com/SteveKGYang/SCCL
\end{abstract}

}

\maketitle

\IEEEdisplaynontitleabstractindextext

%
\IEEEpeerreviewmaketitle

\IEEEraisesectionheading{\section{Introduction}\label{sec:introduction}}

%
%
%
%
\IEEEPARstart{E}{motion} Recognition in Conversations (ERC) aims at identifying the emotion of each utterance within a dialogue from a pre-defined emotion category set~\cite{8764449}. In recent years, ERC has attracted increasing research interest from the NLP community due to the growing availability of public datasets~\cite{busso2008iemocap,poria-etal-2019-meld,zahiri:18a} and its wide applications. For example, ERC enables dialogue systems to generate emotionally coherent and empathetic responses~\cite{MA202050}. It has also been utilised for opinion mining from customer reviews~\cite{9454192,wang2020sentiment} and emotion-related social media analysis~\cite{nandwani2021review,CHATTERJEE2019309}, etc.

Context modelling is a key challenge for ERC. The emotion of each utterance is influenced both by the previous utterances of the speaker and the responses of other participants~\cite{Shen2021DialogXLAX}. Current methods mainly utilise Pre-trained Language Models (PLMs)~\cite{qiu2020pre} to deal with this challenge. However, PLMs are found to poorly capture the semantic meaning of sentences without careful fine-tuning~\cite{li-etal-2020-sentence}, which also raises difficulties for the identification of semantically similar emotions (e.g., \emph{excited} and \emph{happy}). Since previous works utilise unsupervised contrastive learning to alleviate this problem~\cite{li-etal-2020-sentence,DBLP:conf/acl/GiorgiNWB20} and obtain promising results in several text classification tasks, Li et al.\cite{li2021contrast} manage to introduce Supervised Contrastive Learning (SCL) to ERC, where utterances with the same emotion label are considered as positive pairs, and the instance-level utterance representations are directly utilised for contrastive learning. SCL decouples the overlap between samples with similar emotions in the semantic space, and facilitates the learning of the decision boundary.

However, SCL treats two samples as a negative pair as long as they are with different labels, regardless of the quantitative semantic similarity between emotions (e.g., \emph{happy} is closer to \emph{excited} than \emph{sad}). This negligence is manifested by the fact that the representation similarities between the current sample and all negative samples are minimised at the same rate in standard SCL loss. Besides, the success of works with manual feature selection \cite{alswaidan2020survey} shows that ERC is not equally dependent on all features embedded in the high-dimensional utterance representations. We expect a low-dimensional prototype for each emotion, which is defined as \emph{a representative embedding for a group of similar instances}~\cite{PCL}, to be more efficient in contrastive learning. High-dimensional SCL space also leads to other limitations: (a) The curse of dimensionality~\cite{li2021comatch}. (b) The results are hard to interpret and visualise. (c) Stable SCL requires large batch sizes~\cite{khosla2020supervised} which leads to high computational cost.

\begin{figure}[htpb]
\centering
\includegraphics[width=7cm,height=4.9cm]{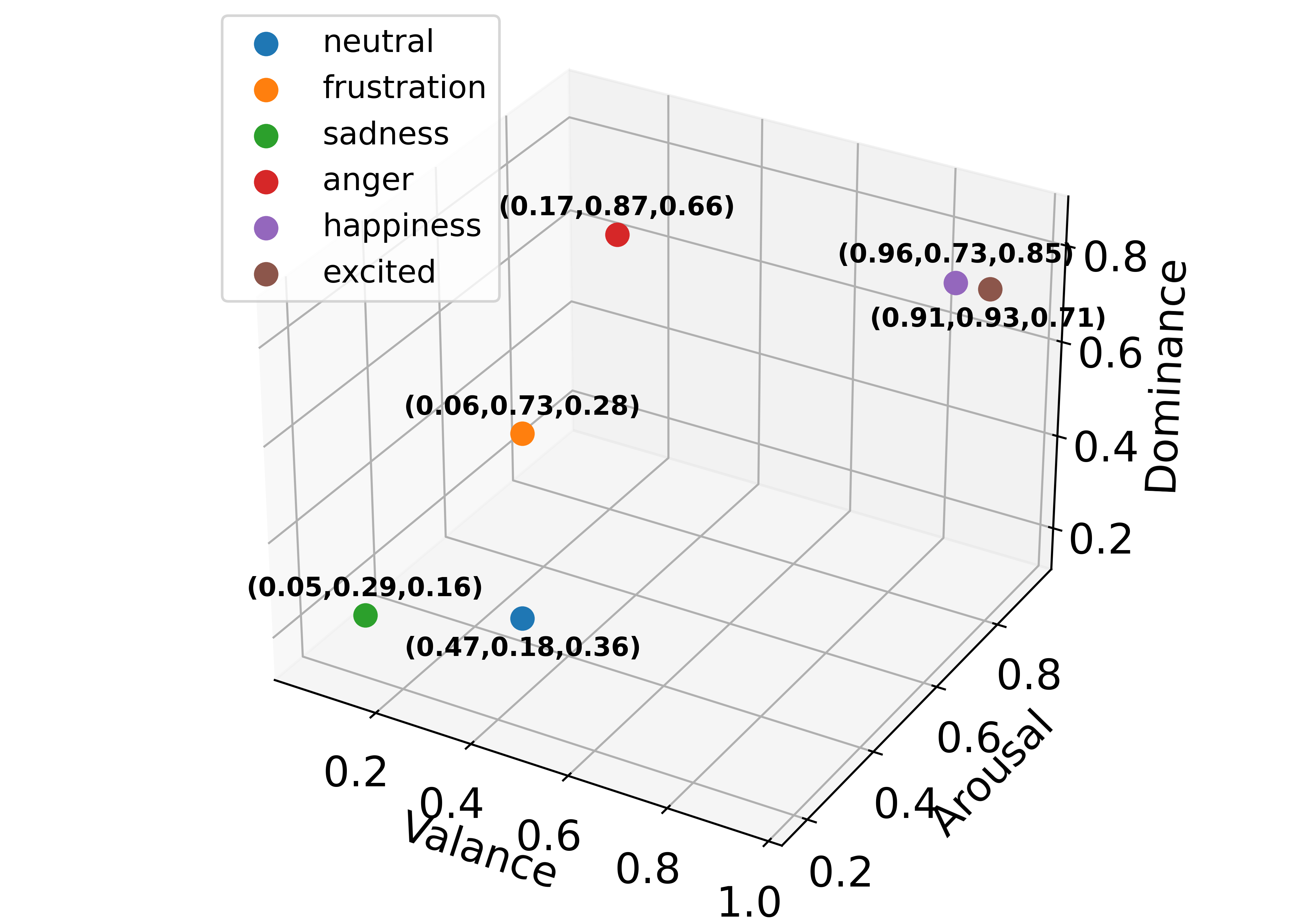}
\caption{An example of appropriate emotion prototypes in VAD space bringing quantitative information.}
\label{fig:example}
\end{figure}

To tackle the above challenges, we propose a novel low-dimensional Supervised Cluster-level Contrastive Learning (SCCL) method for ERC. With a PLM-based context-aware utterance encoder, we improve SCL as follows: (a) we reduce the high-dimensional contrastive learning space to a three-dimensional space called Valence-Arousal-Dominance (VAD), a widely explored affect representation model in psychology \cite{RUSSELL1977273,mehrabian1995framework}. (b) we introduce a human-labelled prototype for each emotion in VAD space, which brings quantitative information between all emotion labels. We provide an example for some emotions in Figure \ref{fig:example}, where the emotions within the same sentiment polarity lie closer and their relative positions are reasonable. Regarding each emotion category as a cluster centre, SCCL predicts the cluster-level VAD for each emotion and transfers the instance-level emotion labels to cluster level with the emotion prototypes, then performs cluster-level contrastive learning. Meanwhile, Liu et al.\cite{liu-etal-2019-linguistic} argue that current PLMs lack fine-grained linguistic knowledge, which is proved useful to help modelling the utterances in sentiment-related tasks~\cite{ke-etal-2020-sentilare}. Factual knowledge is defined as the fact-related commonsense knowledge stored in text-based triplets or sentences~\cite{wang-etal-2021-k}, which is also widely leveraged in ERC~\cite{xie-etal-2021-knowledge-interactive,zhong2019knowledge,ghosal-etal-2020-cosmic} and proved effective in enriching the context and providing relevant knowledge for emotion reasoning. Therefore, we infuse linguistic and factual knowledge utilising the pre-trained knowledge adapter in a plug-in manner, which avoids modification of the PLM weights. Experimental results show that our method achieves state-of-the-art results on three widely used datasets: IEMOCAP, MELD, and DailyDialog. Further analysis shows the effectiveness of each proposed module.

To summarise, this work mainly makes the following contributions:
\begin{itemize}
    \item We reduce the high-dimensional SCL space to a three-dimensional space VAD, which improves model performance and facilitates interpretability.
    \item For the first time in ERC, we incorporate VAD prototypes to SCL by proposing a novel Supervised Cluster-level Contrastive Learning (SCCL). Analysis shows that SCCL remains stable with both large and small batch sizes.
    \item We infuse linguistic and factual knowledge into the utterance encoder by utilising the pre-trained knowledge adapters, and analyse their benefits via the ablation study and empirical comparisons.
\end{itemize}

\begin{figure*}[htpb]
\centering
\includegraphics[width=18cm,height=8cm]{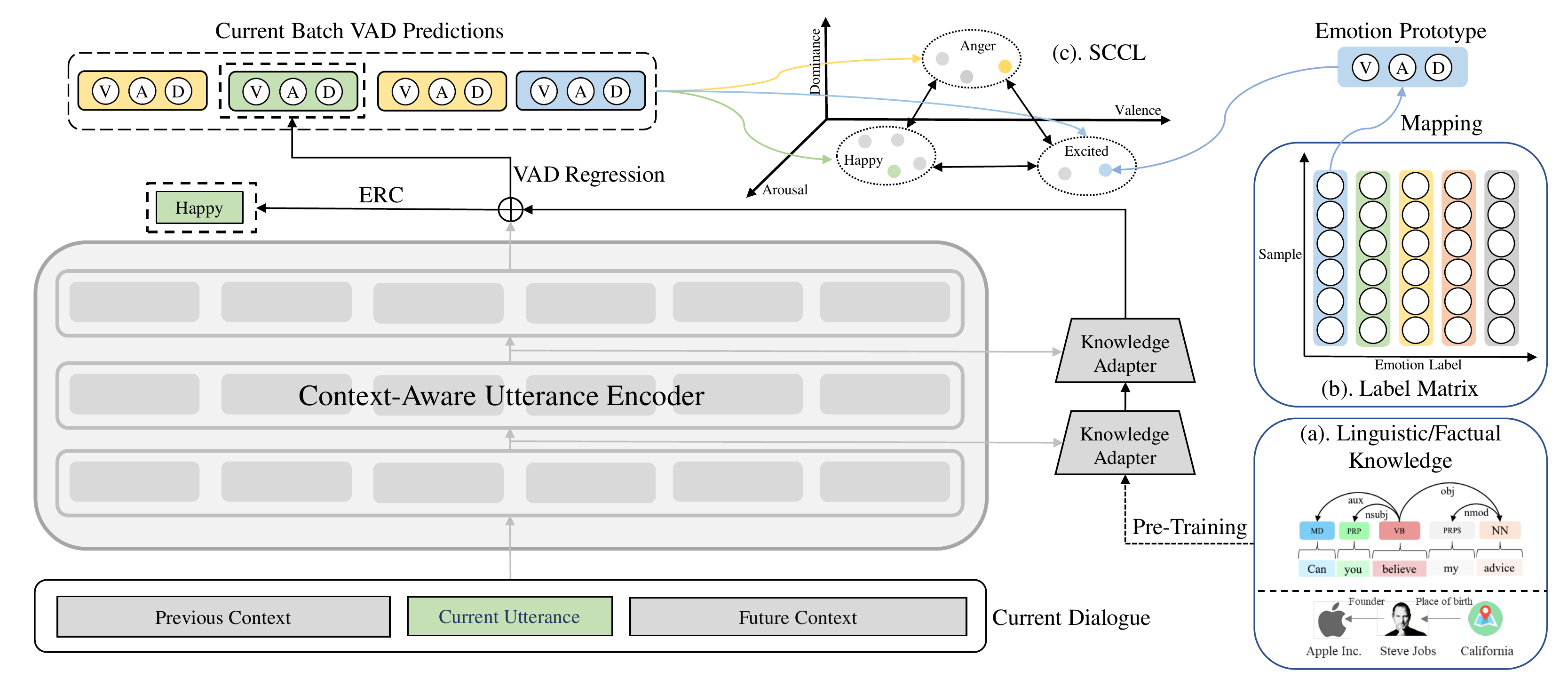}
\caption{An overview of our model architecture. In ERC and SCCL, each colour denotes an emotion category. (a). linguistic and factual knowledge are infused to the knowledge adapters through pre-training, and combined to the utterance encoder in a plug-in manner (Sec. \ref{ki}). (b). The one-hot label matrix is mapped to the VAD space with emotion prototypes (Sec. \ref{sc}). (c). Cluster-level representations are aggregated from the VAD predictions, and contrastive learning is performed in the VAD space (Sec. \ref{sc}).}
\label{fig:model}
\end{figure*}

\section{Related Work}
\subsection{Emotion Recognition in Conversations}
A key challenge of ERC is to leverage rich information in the dialogue context. Early works utilise Recurrent Neural Networks (RNN) to model the intra-speaker dependencies within the utterance sequence for each of the dialogue participants~\cite{hazarika-etal-2018-icon,hazarika-etal-2018-conversational}, and revise the output at each time step as memories. Considering the inter-speaker dependencies, DialogueRNN~\cite{majumder2019dialoguernn} proposes a global state RNN to model multi-party relations and emotional dynamics. Another branch of work leverages the strong context modelling ability of Transformer-based networks to model the dialogue as a whole~\cite{li-etal-2020-hitrans,Shen2021DialogXLAX,kim2021emoberta}. To introduce more interpretable structures, there are also many works~\cite{ghosal-etal-2019-dialoguegcn,shen-etal-2021-directed,liang2021s+} that construct a graph on the dialogue, and devise graph neural networks to model ERC as a node-classification task.

Constrained by the size of available datasets, many works manage to infuse task-related information to aid emotion reasoning. Some methods~\cite{zhang-etal-2020-knowledge,xie-etal-2021-knowledge-interactive,ghosal-etal-2020-cosmic} explicitly incorporate commonsense knowledge to enrich the semantic space. Hazarika et al.\cite{HAZARIKA20211} and Chapuis et al. \cite{chapuis-etal-2020-hierarchical} design relevant pre-training task and transfer the pre-trained weights to ERC. Sentiment scores~\cite{xie-etal-2021-knowledge-interactive}, topic information~\cite{wang2020sentiment,zhu-etal-2021-topic} and speaker-utterance relations~\cite{li-etal-2020-hitrans} are also leveraged to enhance model performance. As an effective dimensional emotion representation model~\cite{buechel2016emotion}, VAD information is also incorporated to facilitate emotion recognition in multiple modalities, such as text~\cite{zhong-etal-2019-knowledge,park-etal-2021-dimensional,10.1145/3404835.3463080} and acoustics~\cite{grimm2007primitives,xia2015multi}, which considerably boosts the model performance.

\subsection{Contrastive Learning}
Unsupervised Contrastive Learning (UCL) aims to construct training samples in an unsupervised manner. In Computer Vision, Chen et al.\cite{chen2020simple} utilise different data augmentation methods to create new samples, and regard samples obtained from the same picture as positive pairs. Based on that idea, Li et al.\cite{li2021contrastive} propose contrastive clustering to produce clustering-favorite representations, which regards each classification class as a cluster, and obtains positive pairs from two different data augmentation methods. Then contrastive learning is performed on both instance and cluster levels. There are also many attempts to reduce the high-dimensional UCL space to incorporate prior knowledge~\cite{PCL,wang2021low}, boost semi-supervised learning~\cite{li2021comatch} and visualise the results~\cite{Zhu_2021_ICCV}. With a similar training framework in NLP, UCL is mainly devised to enforce the sentence representations of PLMs to distinguish similar semantics. For example, Yan et al.\cite{yan-etal-2021-consert} develop data augmentation methods for texts, and Kim et al.\cite{kim-etal-2021-self} train a Siamese model to construct positive pairs.

On the other hand, SCL fully leverages the supervision signals to obtain separable sentence embeddings, which facilitates the model to find the decision boundaries. Existing approaches take samples with the same label as positive pairs to calculate contrastive loss~\cite{gao-etal-2021-simcse,gunel2020supervised,khosla2020supervised}. In emotion recognition, Li et al.\cite{li2021contrast} combine SCL in a multi-task learning setting, which aims to make similar utterances mutually exclusive. Besides, Alhuzali and Ananiadou\cite{9513563} introduces a variant of triplet centre loss that combines both intra- and inter-class variations into the emotion classification loss function.

\section{Methodology}
\subsection{Task Definition and Model Overview}
We define ERC task as follows: a dialogue $D$ in the dataset contains a series of utterances $\{D_1, D_2,..., D_n\}$, with the corresponding emotion labels $\{Y_1, Y_2,..., Y_n\}$, where $Y_i\in E$ is a discrete value indicating the emotion label, and $E$ is the pre-defined categorical emotion set. Each utterance $D_i$ consists of $n_i$ tokens, denoted as $D_i=\{D_i^1, D_i^2, ..., D_i^{n_i}\}$. $D_i$ is uttered by $P(D_i)$, where $P(D_i)\in P$ denotes the speaker, and $P$ is the set of dialogue participants' names. Given the above information, ERC aims to identify the emotions of each utterance, which can be formalized as $\hat{Y}_i=f(D_i, D, P(D_i))$. The overview of our model is presented in Figure \ref{fig:model}.

\subsection{Context-Aware Utterance Encoder}
To introduce speaker information, we pre-pend the name of the speaker $P(D_j)$ for each utterance $D_j$ as $\hat{D}_j$. Then the current utterance $\hat{D}_i$ is concatenated with both past and future contexts to get the context-aware input $R_i$: $R_i = \{[CLS];\hat{D}_{i-W_p};...;\hat{D}_i;...;\hat{D}_{i+W_f};[EOS]\}$, where $W_p$ and $W_f$ denotes past and future context window size, $[CLS]$ and $[EOS]$ denote the start-of-sentence and end-of-sentence token in PLMs. Then we use $R_i$ to obtain the context-aware utterance embeddings:

\begin{equation}
    H_i^L = Encoder(R_i)
\end{equation}

where $Encoder$ denotes the RoBERTa~\cite{liu2019roberta} encoder, $H_i^L\in \mathbb{R}^{S\times D_h}$ denotes the final output of the $L$-th layer, $S$ denotes sequence length and $D_h$ is the hidden size of the encoder.

\subsection{Knowledge Infusion with Adapter}\label{ki}
We incorporate external knowledge into the utterance encoder by injecting pre-trained knowledge adapters. The knowledge adapter is a multi-layer Transformer-based model separately initialised and pre-trained for each of the knowledge sources. During pre-training, the weights of the PLM are fused and only the knowledge adapter weights are updated. Compared with normal pre-training or explicit incorporation methods of knowledge infusion, this training paradigm has three advantages: (1). The weight fusion prevents the catastrophic forgetting~\cite{doi:10.1073/pnas.1611835114} problem of PLMs when multiple knowledge sources are infused. (2). The training process saves memory and speeds up since the knowledge adapter is smaller in size than the PLM. (3). With a new knowledge source to incorporate, the weights of the PLM do not need retraining. 

As shown in Figure \ref{fig:model}(a), we follow the methodology of Wang et al.~\cite{wang-etal-2021-k} and pre-train two knowledge adapters with commonsense knowledge from T-REx~\cite{elsahar-etal-2018-rex} (FacAdapter) and linguistic knowledge provided by Stanford Parser\footnote{https://nlp.stanford.edu/software/lex-parser.html} (LinAdapter). T-REx is a large-scale factual knowledge graph built from over 11.1M alignments between statements and triples of Wikipedia, which provide relevant knowledge to enrich the context and aid emotion reasoning. For example, the statement "Vincent van Gogh and other late 19th century painters used blue not just to depict nature, but to create bad moods and emotions" is aligned with triples $<$Vincent van Gogh, occupation, painters$>$ and $<$blue, represent, bad moods and emotions$>$. During the pre-training process, given the statements and entities as input, the FacAdapter predicts the relation type of the aligned triples. Linguistic knowledge is naturally embedded in language texts, which benefits sentence modelling. It can be obtained by running dependency parser to get semantic and syntactic information. Therefore, for pre-training on linguistic knowledge, the LinAdapter takes the texts as input and predicts the syntactic and semantic relations annotated by the parser.

The adapter is combined as follows: let $Encoder^l$ denotes the $l$-th hidden layer of the utterance encoder. LinAdapter, denoted as \emph{Adapter}, has $n_k$ Transformer-based layers, where $n_k\leq L$ and \emph{Adapter}$^j$ denotes $j$-th layer of the adapter. \emph{Adapter} is also pre-defined an interactive layer set $\hat{L} = \{l_1, l_2,...,l_{n_k}\}$, where the hidden states of $Encoder^{l_j}$ will be combined in \emph{Adapter}$^j$. Specifically, for $i$-th utterance and each $l_j\in \hat{L}$, this process can be formalised as:
\begin{equation}
H_f^j = H_i^{l_j} \oplus H_a^{j-1}\label{eqn4}
\end{equation}\vspace{-.5cm}
\begin{equation}
H_a^{j} = \emph{Adapter}^{j}(H_f^j)\label{eqn5}
\end{equation}

where $H_a^{j}\in \mathbb{R}^{D_h}$ denotes the $j$-th layer output of the knowledge adapter, $H_i^{l_j}$ is the $l_j$-th layer output of the utterance encoder, $\oplus$ denotes element-wise addition, and $H_a^{1}$ is initialised with an all-zero matrix. The final layer output $H_a^{n_k}$ of the adapter is combined with the PLM embeddings as the final utterance representations:
\begin{equation}
\hat{H}_i = Tanh((H_i^{L} \oplus H_a^{n_k})W_1+b_1)\label{eqn6}
\end{equation}

where $\hat{H}_i\in \mathbb{R}^{S\times D_h}$ denotes the knowledge-enhanced utterance embeddings, $Tanh$ denotes the tanh activation function, and $W_1\in \mathbb{R}^{D_h\times D_h}$, $b_1\in \mathbb{R}^{D_h}$ are learnable parameters.

\subsection{Supervised Cluster-Level Contrastive Learning}\label{sc}
\subsubsection{Emotion Prototypes}
Valence-Arousal-Dominance (VAD) maps emotion states to a three-dimensional continuous space, where Valence reflects the pleasantness of a stimulus, arousal reflects the intensity of emotion provoked by a stimulus, and dominance reflects the degree of control exerted by a stimulus~\cite{warriner2013norms}. Instead of directly leveraging the one-hot categorical emotion labels for supervision, VAD allows each categorical emotion to be projected into the space with measurable distances. A few ERC resources~\cite{busso2008iemocap} are human-labelled with a context-dependent VAD score for each utterance $j$: $\emph{H-VAD}_j\in \mathbb{R}^3$, which can be leveraged for accurately computing emotion prototypes. 

However, utterance-level VAD labels are expensive and unavailable in most cases. For application in more scenarios, we consider the context-independent word-level VAD information from sentiment lexicons. We utilise NRC-VAD~\cite{mohammad-2018-obtaining}, a VAD sentiment lexicon that contains reliable human-ratings of VAD for 20,000 English words. All the terms in NRC-VAD denote or connote emotions, and are selected from commonly used sentiment lexicons and tweets. Each of these terms is first strictly annotated via best-worst scaling with crowdsourcing annotators. Then an aggregation process calculates the VAD for each term ranging from 0 to 1. With the pre-defined categorical emotion set $E$, we extract the VAD for each of the emotion $e\in E$ from NRC-VAD: $\emph{NRC-VAD}_{e}\in \mathbb{R}^{3}$. For example, the emotion \emph{happiness} is assigned: $[0.9600, 0.7320, 0.8500]$. The VAD information from either of the above methods is utilised to obtain cluster-level emotion representations. We expect utterance-level H-VADs to outperform word-level NRC-VADs since they are context-dependent and bear more fine-grained VAD information.

\subsubsection{Cluster-Level Contrastive Learning}
Though VAD prototypes provide useful quantitative information, they are difficult to be infused into SCL. Therefore, we propose to perform SCL at cluster level instead of instance level with a novel SCCL method. Regarding each emotion category as a cluster centre, we perform SCCL with cluster-level representations separately obtained from emotion labels and model predictions, where both processes are introduced below.

We first compute for emotion labels using the emotion prototypes. For a batch $B$, as shown in Figure \ref{fig:model}(b), the emotion labels are projected to a one-hot label matrix $M\in \mathbb{R}^{|B|\times|E|}$, where $M_i\in \mathbb{R}^{|E|}$ is the $i$-th row of $M$, denoting the one-hot emotion label of the $i$-th sample, and $M^j\in \mathbb{R}^{|B|}$ is the $j$-th column of $M$, denoting the samples with the label $e_j\in E$. For the $j$-th cluster $e_j$, we map $M^j$ to the VAD space as follows:
\begin{equation}
\hat{M}^j = \frac{\sum_{k=1}^BM^{jk}\times\emph{VAD}_{e_j}}{\sum_{k=1}^BM^{jk}}\label{eqn8}
\end{equation}

where $M^{jk}$ denotes $k$-th element of $M^j$, $\hat{M}^j\in \mathbb{R}^3$ denotes the cluster-level representation of $e_j$. When utterance-level VAD information is available, $\emph{VAD}_{e_j}=\emph{H-VAD}_j$. When NRC-VAD information is utilised, $\emph{VAD}_{e_j}=\emph{NRC-VAD}_{e_j}$ and $\emph{NRC-VAD}_{e_j}$ is directly regarded as the cluster-level emotion representation for $e_j$. 

Then we compute for the model predictions. One choice is to adopt a similar approach as the emotion labels, which utilises the normalised categorical predictions with Softmax, and maps them to the VAD space using Eqn.\ref{eqn8}. However, it may reduce SCCL to the vanilla case where the model only learns the one-hot label information and ignores the emotion prototypes. Therefore, we utilise a neural network to parameterise the dimension reduction process from the semantic space to the VAD space. Specifically, for $\hat{H}_i$, we regard the embedding of the start-of-sentence token at position 0 $\hat{H}_i^{[CLS]}$ as its utterance-level embedding, and map $\hat{H}_i^{[CLS]}$ to the VAD space:
\begin{equation}
H_i^{VAD} = \frac{1}{1+\emph{e}^{-(\hat{H}_i^{[CLS]}W_2+b_2)}}\label{eqn10}
\end{equation}

where $\hat{H}_i^{[CLS]}\in\mathbb{R}^{D_h}$, $H_i^{VAD}\in\mathbb{R}^{3}$, and $W_2\in \mathbb{R}^{D_h\times 3}$, $b_2\in \mathbb{R}^{3}$ are learnable parameters. 

As shown in Figure \ref{fig:model}(c), following the idea of \emph{labels as representations}, for each batch, we calculate the SCCL loss as follows:
\begin{equation}
    \hat{H}^{VAD}_j=\frac{1}{|\mathcal{A}(j)|}\sum_{i\in\mathcal{A}(j)}H^{VAD}_i\label{eqn11}
\end{equation}
\begin{equation}
    sim(j)=log\frac{exp(\hat{H}^{VAD}_j\cdot \hat{M}^j)/\tau}{\sum_{e_k\in E} exp(\hat{H}^{VAD}_j\cdot \hat{M}^k)/\tau}
\end{equation}
\begin{equation}
    \mathcal{L}_{SCCL} = -\frac{1}{|E|}\sum_{e_j\in E}sim(j)
\end{equation}

where $\hat{H}^{VAD}_j\in \mathbb{R}^3$ denotes the cluster-level embedding for $e_j$ from model predictions, $\mathcal{A}(j)=\{i|Y_i=e_j,i\in [1,|B|]\}$ records the samples $D_i\in B$ labelled with the emotion $e_j$, $\cdot$ denotes dot-product operation, $\tau\in\mathbb{R}^{+}$ is the temperature coefficient, and $\mathcal{L}_{SCCL}$ denotes the SCCL loss.

\subsection{Model Training}
We combine SCCL with ERC in a multi-task learning manner. For the $i$-th utterance, we still utilise $\hat{H}_i^{[CLS]}$ as the utterance-level embedding, and compute the final classification probability as follows:
\begin{equation}
    \hat{Y}_i = Softmax(\hat{H}_i^{[CLS]}W_3+b_3)
\end{equation}

where $\hat{Y}_i\in \mathbb{R}^{|E|}$, and $W_3\in \mathbb{R}^{D_h\times |E|}$, $b_3\in \mathbb{R}^{|E|}$ are learnable parameters. Then we compute the ERC loss using the standard cross-entropy loss:
\begin{equation}
\begin{aligned}
    \mathcal{L}_{ERC}=-\frac{1}{B}\sum_{i=1}^B\sum_{j=1}^{|E|}Y_i^jlog\hat{Y}_i^j
\end{aligned}
\end{equation}

where $Y_i^j$ and $\hat{Y}_i^j$ are the $j$-th element of $Y_i$ and $\hat{Y}_i$. Finally, we combine the ERC loss and SCCL loss in the following manner:
\begin{equation}\label{eqn12}
    \mathcal{L} = \mathcal{L}_{ERC}+\alpha \mathcal{L}_{SCCL}
\end{equation}

where $\alpha\in [0, 1]$ denotes the pre-defined weight coefficient of $\mathcal{L}_{SCCL}$.

\section{Experimental Settings}
\subsection{Datasets}
We evaluate our method on the following four benchmark datasets. The statistics of all datasets are presented in Table \ref{tab:statistics}.

\begin{table}[h]
\caption{Statistics of the datasets. Conv. and Utter. denotes the conversation and utterance number. Utter./Conv denotes the average utterance number per dialogue.}
\label{tab:statistics}
\begin{center}
\setlength{\tabcolsep}{0.5mm}{
\begin{tabular}{lccc}
\toprule \textbf{Dataset} &  \textbf{Conv.(Train/Val/Test)} & \textbf{Utter.(Train/Val/Test)} &  \textbf{Utter./Conv} \\ \midrule
IEMOCAP & 100/20/31 &  4,778/980/1,622 & 49.2 \\
MELD &  1,038/114/280 & 9,989/1,109/2,610 & 9.6\\
EmoryNLP &  713/99/85 & 9,934/1,344/1,328 & 14.1 \\
DailyDialog & 11,118/1,000/1,000 & 87,170/8,069/7,740 & 7.9 \\
\bottomrule
\end{tabular}}
\end{center}
\end{table}

\textbf{IEMOCAP}~\cite{busso2008iemocap}: A two-party multi-modal conversation dataset derived from the scenarios in the scripts of the two actors. For all datatsets, we only utilise the text modality in our experiments. The pre-defined categorical emotions are \emph{neutral, sad, anger, happy, frustrated, excited}.

\textbf{MELD}~\cite{poria-etal-2019-meld}: A multi-party multi-modal dataset enriched from \emph{EmotionLines} dataset, collected from the scripts of American TV show \emph{Friends}. The pre-defined emotions are \emph{neutral, sad, anger, disgust, fear, happy, surprise}.

\textbf{EmoryNLP}~\cite{zahiri:18a}: Another dataset collected from TV show \emph{Friends}, but annotated with different emotion label categories. The pre-defined emotions are \emph{neutral, sad, mad, scared, powerful, peaceful, joyful}.

\textbf{DailyDialog}~\cite{li-etal-2017-dailydialog}: A dataset compiled from human-written daily conversations with only two parties involved and no speaker information. The pre-defined emotion labels are the Ekman’s emotion types: \emph{neutral, happy, surprise, sad, anger, disgust, fear}.

\begin{table}[h]
\caption{The NRC-VAD assignments to all emotions in the four datasets.}
\label{vads}
\begin{center}
\resizebox{.5\textwidth}{!}{
\begin{tabular}{l|ccccccc}
\toprule \textbf{IEMOCAP} & neutral & frustrated & sad & anger & excited & happy & --\\ \midrule
Valence & 0.469 & 0.060 & 0.052 & 0.167 & 0.908 & 0.960 & --\\
Arousal & 0.184 & 0.730 & 0.288 & 0.865 & 0.931 & 0.732 & --\\
Dominance & 0.357 & 0.280 & 0.164 & 0.657 & 0.709 & 0.850 & --\\
\midrule\midrule
\textbf{MELD} & neutral & joy & surprise & anger & sad & disgust & fear\\ \midrule
Valence & 0.469 & 0.980 & 0.875 & 0.167 & 0.052 & 0.052 & 0.073\\
Arousal & 0.184 & 0.824 & 0.875 & 0.865 & 0.288 & 0.775 & 0.840\\
Dominance & 0.357 & 0.794 & 0.562 & 0.657 & 0.164 & 0.317 & 0.293\\
\midrule\midrule
\textbf{EmoryNLP} & joyful & neutral & powerful & mad & sad & scared & peaceful\\ \midrule
Valence & 0.990 & 0.469 & 0.865 & 0.219 & 0.225 & 0.146 & 0.867\\
Arousal & 0.740 & 0.184 & 0.830 & 0.873 & 0.333 & 0.828 & 0.108\\
Dominance & 0.667 & 0.357 & 0.991 & 0.277 & 0.149 & 0.185 & 0.569\\
\midrule\midrule
\textbf{DailyDialog} & neutral & anger & disgust & fear & happy & sad & surprise\\ \midrule
Valence & 0.469 & 0.167 & 0.052 & 0.073 & 0.960 & 0.052 & 0.875\\
Arousal & 0.184 & 0.865 & 0.775 & 0.840 & 0.732 & 0.288 & 0.875\\
Dominance & 0.357 & 0.657 & 0.317 & 0.293 & 0.850 & 0.164 & 0.562\\
\bottomrule
\end{tabular}}
\end{center}
\end{table}

Among the above datasets, human-labelled utterance-level VAD scores are only available in IEMOCAP, where the aggregation process calculates the VAD for each utterance ranging from 1 to 5. To cope with the SCCL method, we linearly transform all VAD scores to the range $[0,1]$ during inference.

When NRC-VAD is utilised, the emotion prototypes of the labels for all datasets are listed in Table \ref{vads}. According to the assignments, most of the cluster centres reflect appropriate positions of the corresponding emotions in VAD space, where similar emotions are measurably closer to each other while maintaining a fine-grained difference to facilitate the model to distinguish them. For example, \emph{happy} stays closer to \emph{excited} than \emph{anger} in IEMOCAP. In addition, for all four datasets, positive and negative emotions are mostly separated by \emph{neutral} in the dimension Valence, while the emotions within each sentiment polarity mostly differs in Arousal and Dominance. 

\subsection{Baselines}
We select the following strong baseline models to compare with our model:

\textbf{BERT-Large}~\cite{devlin-etal-2019-bert}: Initialised from pre-trained weights of BERT-Large. The $[CLS]$ embedding is fine-tuned for ERC.

\textbf{DialogXL}~\cite{Shen2021DialogXLAX}: This PLM-based work uses dialog-aware self-attention to model inter- and intra-speaker dependencies, and utilises utterance recurrence to model long-range contexts.

\textbf{RGAT}~\cite{ishiwatari-etal-2020-relation}: The model constructs a graph to introduce prior knowledge in context modelling, and combines a relation position encoding to introduce sequential information.

\textbf{COSMIC}~\cite{ghosal-etal-2020-cosmic}: Based on RNN structure to model dependencies, this work introduces utterance-level commonsense knowledge to model the mental states of speakers.

\textbf{KI-Net}~\cite{xie-etal-2021-knowledge-interactive}: This work leverages token-level commonsense knowledge from knowledge graphs to enrich contexts, and implicitly introduces sentiment scores from sentiment lexicons to guide emotion reasoning.

\textbf{DAG-ERC}~\cite{shen-etal-2021-directed}: Based on RoBERTa-Large, this model builds a Directed Acyclic Graph (DAG) on the dialogue. Then DAGNN is used for graph embedding.

\textbf{SGED}~\cite{bao2022speaker}: This method proposes a speaker-guided encoder-decoder framework to exploit speaker information for ERC.

\textbf{SKAIG}~\cite{li-etal-2021-past-present}: This work builds a graph on the dialogue and utilises psychological commonsense knowledge to enrich edge representations. 

\textbf{CoG-BART}~\cite{li2021contrast}: Based on BART-Large~\cite{lewis-etal-2020-bart}, this work utilises SCL and a response generation auxiliary task to distinguish semantics of utterances with similar emotions.

\subsection{Implementation Details}
We conduct the experiments on IEMOCAP, MELD, and EmoryNLP using a single Nvidia Tesla V100 GPU with 16GB of memory, and set the batch size to 4. For large-scale dataset DailyDialog, we conduct the experiments using a single Nvidia Tesla A100 GPU with 80GB of memory, and set the batch size to 16. We initialise the pre-trained weights of PLMs and use the tokenization tools both provided by Huggingface\footnote{\url{https://huggingface.co/}}. The pre-trained knowledge adapter weights are from Wang et al.\cite{wang-etal-2021-k}, and these weights are fused during training. We leverage AdamW optimiser~\cite{DBLP:conf/iclr/LoshchilovH19} to train the model, with a linear warm-up learning rate scheduling~\cite{goyal2017accurate} of warm-up ratio 20$\%$ and peak learning rate 1e-5. Due to the limitation of computation memory, we use mixed floating point precision~\cite{micikevicius2017mixed} during training. Hyper-parameters are tuned on the validation set, where $\tau=1.0$ and $\alpha$ is tuned on the interval $[0.5, 1.0]$ and set to 1.0 for MELD and 0.8 for all other datasets. $S=512$, $D_h=1024$, $L=24$ for RoBERTa-Large, and $D_h=768$, $L=12$ for RoBERTa-Base. We set a dropout rate 0.1, a $L^2$ regularisation rate 0.01 to avoid over-fitting. We use the weighted-F1 measure as the evaluation metric for IEMOCAP, MELD and EmoryNLP. Since \emph{neutral} occupies most of DailyDialog, we use micro-F1 for this dataset, and ignore the label \emph{neutral} when calculating the results as in the previous works~\cite{shen-etal-2021-directed,li2021contrast}. All reported results are averages of five random runs.

\section{Results and Analysis}
\subsection{Overall Performance}\label{overall}

Table \ref{tab:results} presents the performance of our method, and compares it to the strong baseline models. 
\begin{table}[h]
\caption{The test results on IEMOCAP, MELD, EmoryNLP and DailyDialog datasets. HVAD-SCCL denotes our SCCL method utilising the utterance-level VAD labels, and NRC-SCCL denotes SCCL with the NRC-VAD supervision signals. All SCCL results are with LinAdapter. Best values are highlighted in bold. The numbers with $*$ indicate that the improvement of our model over all baselines is statistically significant with p $<$ 0.05 under t-test.}
\label{tab:results}
\begin{center}
\setlength{\tabcolsep}{0.5mm}{
\begin{tabular}{l|cccc}
\toprule Model & IEMOCAP & MELD & EmoryNLP & DailyDialog \\ \midrule
BERT-Large~\cite{devlin-etal-2019-bert} & 60.60 & 62.83 & 33.73 & 54.09 \\
DialogXL~\cite{Shen2021DialogXLAX} & 65.94 & 62.41 & 34.73 & 54.93\\ \midrule
COSMIC~\cite{ghosal-etal-2020-cosmic} & 65.28 & 65.21 & 38.11 & 58.48\\
KI-Net~\cite{xie-etal-2021-knowledge-interactive} & 66.98 & 63.24 & -- & 57.30\\
SGED~\cite{bao2022speaker} & 68.53 & 65.46 & \textbf{40.24} & --\\
SKAIG~\cite{li-etal-2021-past-present} & 66.96 & 65.18 & 38.88 & 59.75\\ \midrule
RGAT~\cite{ishiwatari-etal-2020-relation} & 65.22 & 60.91 & 34.42 & 54.31\\
DAG-ERC~\cite{shen-etal-2021-directed} & 68.03 & 63.65 & 39.02 & 59.33\\
CoG-BART~\cite{li2021contrast} & 66.18 & 64.81 & 39.04 & 56.29\\
\midrule
HVAD-SCCL & \textbf{69.88}*($\pm$0.50) & -- & -- & --\\
NRC-SCCL & 69.81*($\pm$0.63) & \textbf{65.70}*($\pm$0.82) & 38.75($\pm$0.49) & \textbf{62.51}*($\pm$0.20)\\
\bottomrule
\end{tabular}}
\end{center}
\end{table}

According to the results, BERT-Large and DialogXL achieves competitive results on all datasets. These PLM-based methods are usually used as foundations for other works. KI-Net, COSMIC and SKAIG all explicitly incorporate factual knowledge at the token level or mental state knowledge at the utterance level, and achieve competitive performance especially on short-context datasets, such as over $57\%$ on DailyDialog (7.9
utterances per dialogue). SGED also implicitly models speaker information via an encoder-decoder framework, which leads to a balanced improvement on all datasets and the best performance 40.24$\%$ on EmoryNLP. These results demonstrate that infusing task-related knowledge and information is beneficial for ERC task. Though RGAT and DAG-ERC both utilise graph structure to model the context, DAG-ERC significantly outperforms RGAT with over 3$\%$ gain on all datasets, showing the importance of more reasonable dialogue modelling structures. The competitive performance of CoG-BART also shows the effectiveness of other representation learning techniques such as supervised contrastive learning and response generation. 

\begin{figure}[htpb]
\centering
\includegraphics[width=6cm,height=6cm]{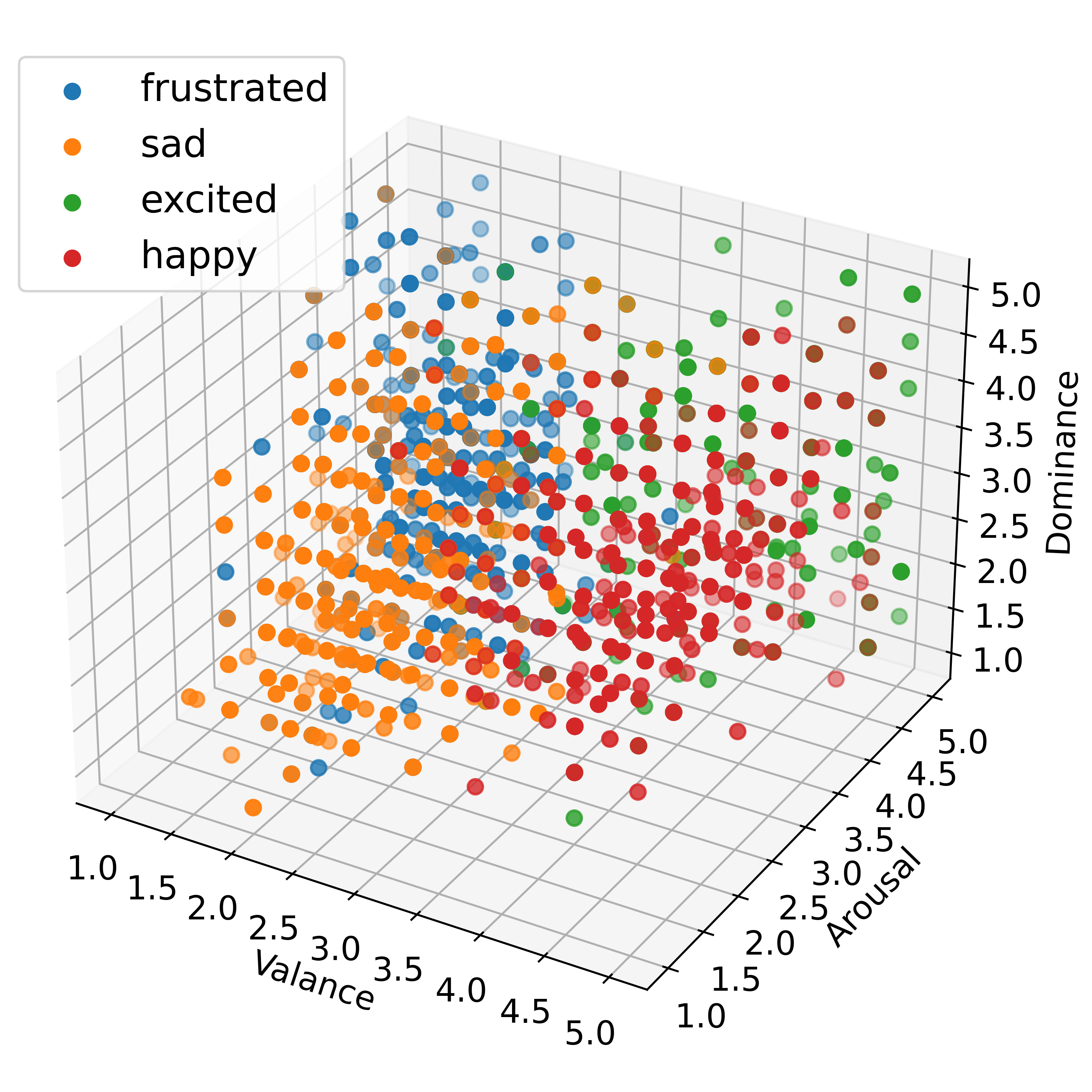}
\caption{Visualisation of HVAD annotations in IEMOCAP training set.}
\label{fig:hvad}
\end{figure}

We can only test HVAD-SCCL on IEMOCAP since all other datasets do not provide human-labelled utterance-level VAD scores. According to the results, HVAD-SCCL achieves a new state-of-the-art result of 69.88\%, but outperforms NRC-SCCL slightly on IEMOCAP, which does not correspond with our early hypothesis. We notice that NRC-VAD follows strict best-worst scaling annotation and aggregation processes with a minimum of 6 annotators per word. In comparison, the IEMOCAP VAD (HVAD) annotation process follows a rough scheme, which brings inaccuracy to the annotated labels and only provides coarse-grained VAD information within each emotion. According to the visualisation results in Figure \ref{fig:hvad}, the VAD shifts within each emotion are mostly discrete, which leads to a limited advantage over the fixed NRC-VAD prototypes. In addition, the VAD distributions of semantically similar emotions (e.g. \emph{Frustrated} and \emph{Sad}) appear to be more entangled, which increases confusion during the training process.

\begin{table*}[h]
\begin{center}
\caption{Some samples of the fuzzy emotion \emph{peaceful} and \emph{powerful} that shift in Valence-Arousal-Dominance. We provide the NRC-VAD emotion prototypes for the two emotions, and the VAD predictions of each utterance in a random run of Lin-SCCL.}
\label{tab:emoryexample}
\setlength{\tabcolsep}{0.3mm}{
\begin{tabular}{l|l}
\toprule
Emotion &  Utterances\\ \midrule
\multirow{6}{*}{\emph{peaceful}\ \textbf{{(0.867,0.108,0.569)}}} & \begin{tabular}[c]{@{}l@{}} 1. Well...you never know. How's. um... how's the family?\ \textbf{(0.460,0.249,0.355)}\end{tabular}\\ & \begin{tabular}[c]{@{}l@{}} 2. Warden, in five minutes my pain will be over. But you'll have to live with the knowledge that you sent an honest\\ man to die.\ \textbf{(0.317,0.470,0.261)}\end{tabular}\\ & \begin{tabular}[c]{@{}l@{}} 3. Yeah, I'm sorry too. But, I gotta tell you. I'm a little relieved. \textbf{(0.752,0.696,0.410)}\end{tabular}\\ & \begin{tabular}[c]{@{}l@{}} 4. Oh, like you've never gotten a little rambunctious with Ross.\ \textbf{(0.341,0.527,0.497)}\end{tabular}\\ & \begin{tabular}[c]{@{}l@{}} 5. Yeah, I'm thinking, if we put our heads together, between the two of us, we can break them up.\ \textbf{(0.770,0.194,0.712)}\end{tabular}\\\midrule
\multirow{8}{*}{\emph{powerful}\ \textbf{(0.865,0.830,0.991)}} & \begin{tabular}[c]{@{}l@{}} 1. ..Dammit, hire the girl! Okay, everybody ready?\ \textbf{(0.483,0.936,0.898)}\end{tabular}\\ & \begin{tabular}[c]{@{}l@{}} 2. Okay, everybody, we'd like to get this in one take, please. Let's roll it... water's working... and...\\ action.\ \textbf{(0.640,0.794,0.519)} \end{tabular}\\ & \begin{tabular}[c]{@{}l@{}} 3. I'm on top of the world, looking down on creation and the only explanation I can find, is the\\ wonders I've found ever since...\ \textbf{(0.722,0.584,0.716)}\end{tabular}\\ & \begin{tabular}[c]{@{}l@{}} 4. Alright, I looked all over the building and I couldn't find the kitty anywhere.\ \textbf{(0.325,0.764,0.372)}\end{tabular}\\ & \begin{tabular}[c]{@{}l@{}} 5. My God, you're choking! That better?\ \textbf{(0.322,0.905,0.569)}\end{tabular}\\\bottomrule
\end{tabular}}
\end{center}
\end{table*}

On the other hand, NRC-SCCL obtains competitive results on all datasets, and achieves new state-of-the-art results 69.81$\%$ on IEMOCAP, 65.70$\%$ on MELD and 62.51$\%$ on DailyDialog. Specifically, NRC-SCCL outperforms all information infusion-based models on three datasets with linguistic knowledge and NRC-VAD emotion prototypes, showing the effectiveness of these information. It also improves the performance of CoG-BART by over 3$\%$ on IEMOCAP and 6$\%$ on DailyDialog, indicating the advantage of SCCL over vanilla supervised contrastive learning and response generation. However, on EmoryNLP, NRC-SCCL fails to outperform the baseline models as in the other datasets. A possible reason is that EmoryNLP defines fuzzy emotions \emph{powerful} and \emph{peaceful}. Though appearing highly positive in NRC-VAD (\emph{Powerful}: [0.865,0.830,0.991], \emph{peaceful}: [0.867,0.108,0.569], as listed in Table \ref{vads}), we find that many utterances labelled with these emotions do not yield positive sentiments. Therefore, unified VAD prototypes of the fuzzy emotions are misleading for many samples. 

We provide some cases in Table \ref{tab:emoryexample} to explain the above hypothesis. All examples are from the training set of EmoryNLP. In the samples of \emph{peaceful}, utterance \#1 expresses no apparent emotions with a moderate Valence score 0.460, utterance \#2 conveys weak sadness and anger with lower Valence 0.317 and higher Dominance 0.470, and utterance \#3 shows implicit happiness with higher Valence 0.752. In the samples of \emph{powerful}, though all utterances express high Arousal (emotion intensity) which corresponds to the NRC-VAD emotion prototypes, these examples have different Valence and Dominance levels. For example, utterance \#1 has high dominance 0.898 with a strong sense of control, but utterance \#2 and \#4 show relatively low dominance 0.519 and 0.372. On the other hand, utterance \#3 conveys high Valence 0.722 with apparent pleasantness, while utterance \#4 and \#5 express sadness and fear with low Valence 0.325 and 0.322. Therefore, the model is unable to learn fine-grained shifts in fuzzy emotions with the unified NRC-VAD emotion prototypes. One direction of our future work is leveraging more fine-grained supervision signals to handle the change of situations for fuzzy emotions.

\begin{table}[h]
\caption{Results of ablation study for two knowledge types. Lin-SCCL denotes the SCCL method with LinAdapter and Fac-SCCL is with FacAdapter. Lin-RL denotes Lin-SCCL with $\mathcal{L}_{SCCL}$ in Eqn. \ref{eqn12} replaced by a correlation-based regression loss on the VAD scores, and other settings remain unchanged. All experiments use the NRC-VAD supervision signals. ``-" denotes the removal of one or several modules, and ``Adapter" denotes the adapter module. The values in parentheses indicate the relative change with respect to Lin-SCCL and Fac-SCCL. We omit the repeated results for Fac-SCCL. Best values are highlighted in bold.}
\label{tab:compare}
\begin{center}
\setlength{\tabcolsep}{0.3mm}{
\begin{tabular}{l|cccc}
\toprule Model & IEMOCAP & MELD & EmoryNLP & DailyDialog \\ \midrule
Fac-SCCL & 69.66 & 65.10 & 37.85 & 61.89\\
\ -SCCL & 68.25($\downarrow$1.41) & 64.20($\downarrow$0.90) & 37.10($\downarrow$0.75) & 60.64($\downarrow$1.25)\\ \midrule\midrule
Lin-SCCL & \textbf{69.81} & \textbf{65.70} & \textbf{38.75} & \textbf{62.51}\\
\ -SCCL & 68.21($\downarrow$1.60) & 63.70($\downarrow$2.00) & 38.53($\downarrow$0.22) & 60.38($\downarrow$2.13)\\
\ -Adapter & 69.23($\downarrow$0.58) & 64.72($\downarrow$0.98) & 37.45($\downarrow$1.30) & 61.53($\downarrow$0.22)\\
\ -SCCL,Adapter & 66.52($\downarrow$3.29) & 63.44($\downarrow$2.26) & 36.68($\downarrow$2.07) & 59.32($\downarrow$3.19)\\
Lin-RL & 68.70($\downarrow$1.11) & 64.65($\downarrow$1.05) & 38.12($\downarrow$0.63) & 61.24($\downarrow$1.27)\\
\bottomrule
\end{tabular}}
\end{center}
\end{table}

\subsection{Ablation Study}
\label{ablation}
We investigate the performance of each proposed module via an ablation study on Lin-SCCL in Table \ref{tab:compare}. According to the results, Lin-SCCL outperforms the context-aware utterance encoder by over 3\% on IEMOCAP and DailyDialog, and over 2\% on MELD and EmoryNLP. These improvements show the joint contribution of linguistic knowledge and SCCL. While the removal of either SCCL or LinAdapter decreases the model performance, removing SCCL leads to a more serious decrease in performance, with over a 1.5\% drop for IEMOCAP, MELD and DailyDialog. According to the previous analysis in NRC-VAD emotion prototypes, SCCL is expected to be beneficial in distinguishing similar emotions, which is crucial for ERC and leads to more improvement than LinAdapter on these datasets. On EmoryNLP, LinAdapter benefits model performance more significantly than SCCL since the fuzzy emotions affect the contrastive learning process in VAD space, as analysed in Sec. \ref{overall}. Lin-SCCL outperforms Lin-RL by over 1\% on most datasets, showing SCCL as more appropriate for leveraging VAD information. A possible reason is that regression loss only introduces the current emotion's cluster-level representation, while SCCL also introduces and pushes apart all other emotion prototypes. SCCL further enables the model to be aware of the quantitative information between each pair of emotions.

\subsection{Empirical Comparison of Knowledge Adapters}
We analyse the effect of linguistic knowledge and factual knowledge on SCCL and the utterance encoder by comparing their performance in ERC, where the results are shown in Table \ref{tab:compare}. According to the results, both LinAdapter and FacAdapter contribute to the performance positively, denoting the effectiveness of both knowledge types. Lin-SCCL outperforms Fac-SCCL on all four datasets, because linguistic knowledge provides utterance structure information to help discover the linguistic patterns for emotion expression, which benefits the contrastive learning process. On the other hand, much factual knowledge is unrelated to affect and brings noise to the fine-grained emotion reasoning in SCCL. With the removal of SCCL, the utterance encoder achieves superior results on MELD and DailyDialog with FacAdapter, since the factual knowledge enriches the semantics of utterances, which benefits the dialogues with short contexts. This hypothesis is further indicated by the more significant improvement with LinAdapter on the other two rich-context datasets IEMOCAP and EmoryNLP. Overall, the empirical comparison of both knowledge adapters verifies the more benefits of linguistic knowledge on SCCL, and factual knowledge provides more information to the utterance encoder in short-context scenarios.

\begin{figure}[H]
\centering
\includegraphics[width=7cm,height=8.4cm]{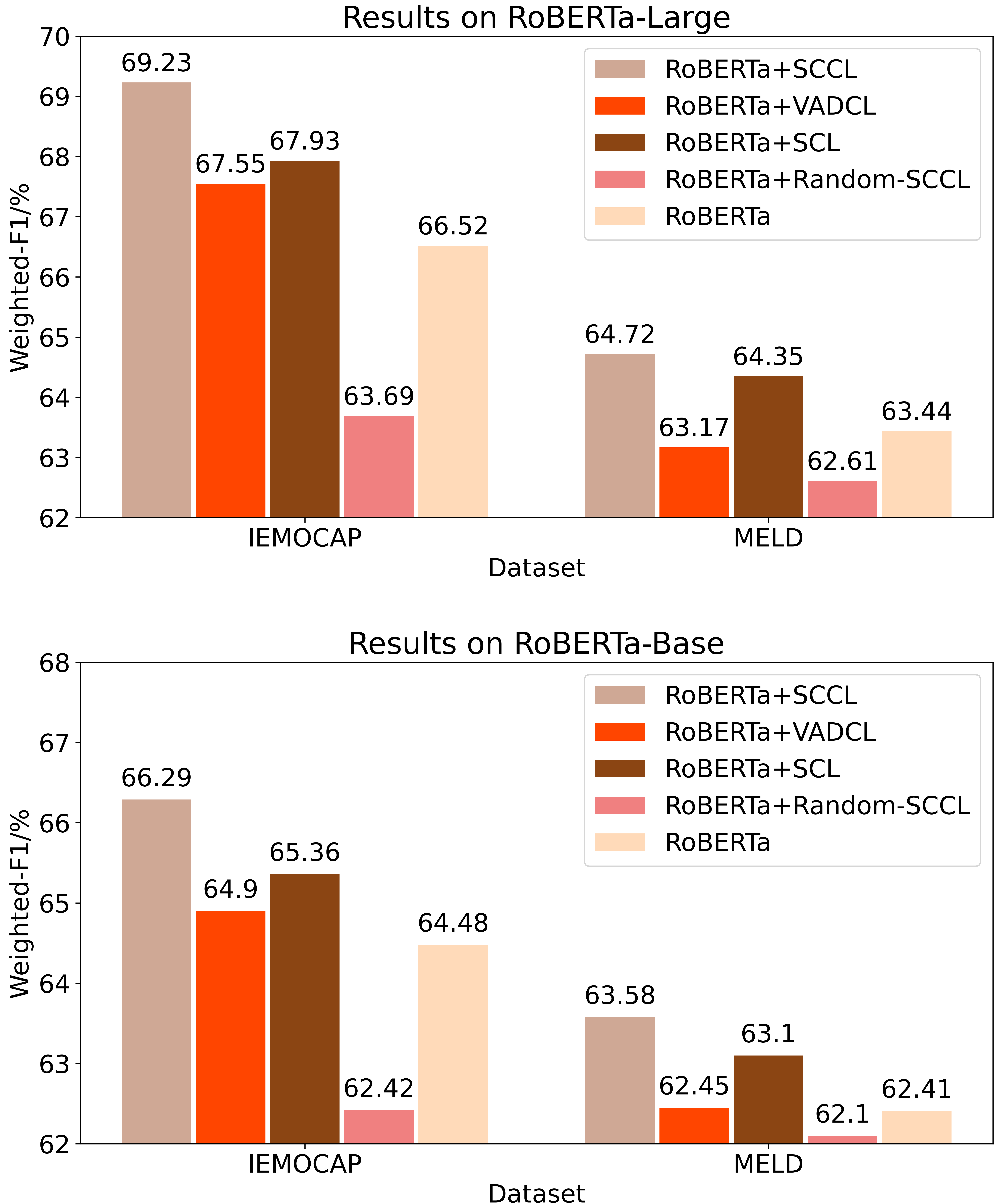}
\caption{Performance of different contrastive learning methods with RoBERTa-Large and RoBERTa-Base encoder.}
\label{fig:bar}
\end{figure}

\subsection{Comparison of Contrastive Learning Methods}
We compare the results of different contrastive learning methods with RoBERTa encoder in Figure \ref{fig:bar}. 

Both large and base-size encoders are leveraged to compare the performance of encoders with different context modelling ability. ``VADCL'' denotes performing SCL directly on VAD space without introducing emotion prototypes, and ``Random-SCCL'' utilises the same structure as SCCL except \emph{randomly initialising} the prototype for each emotion instead of utilising NRC-VAD.

For the results on RoBERTa-Large, SCCL outperforms the RoBERTa baseline with an improvement of 2.71$\%$ on IEMOCAP and 1.28$\%$ on MELD. VADCL achieves comparable performance with SCL on both datasets, proving the viability of performing contrastive learning on a low-dimensional space instead of the semantic space, which also provides useful information to facilitate the identification of emotions. SCCL also outperforms VADCL on both datasets, denoting that emotion prototypes guide samples of each emotion to cluster towards proper positions and maintain appropriate quantitative relations. To further analyse this hypothesis, we experiment on RoBERTa+Random-SCCL, and Random-SCCL yields worse outcomes than RoBERTa on both datasets. These results indicate that SCCL relies on emotion prototypes instead of merely clustering the same emotion as in SCL. The quantitative information embedded in the prototype of each emotion is eliminated as the consequence of the random initialisation, and these false relations mislead SCCL.

We also present the results with RoBERTa-Base encoder. As expected, RoBERTa-Large outperforms RoBERTa-Base with all contrastive learning methods. Similar conclusions about the comparisons of contrastive learning methods are drown from the results of RoBERTa-Base, showing that our above conclusions are robust with utterance representations of varying quality. In addition, the advantage of SCCL is more apparent on IEMOCAP, showing the consistent benefits of rich context on SCCL with different utterance encoders.

\begin{figure}[htpb]
\centering
\includegraphics[width=9cm,height=6cm]{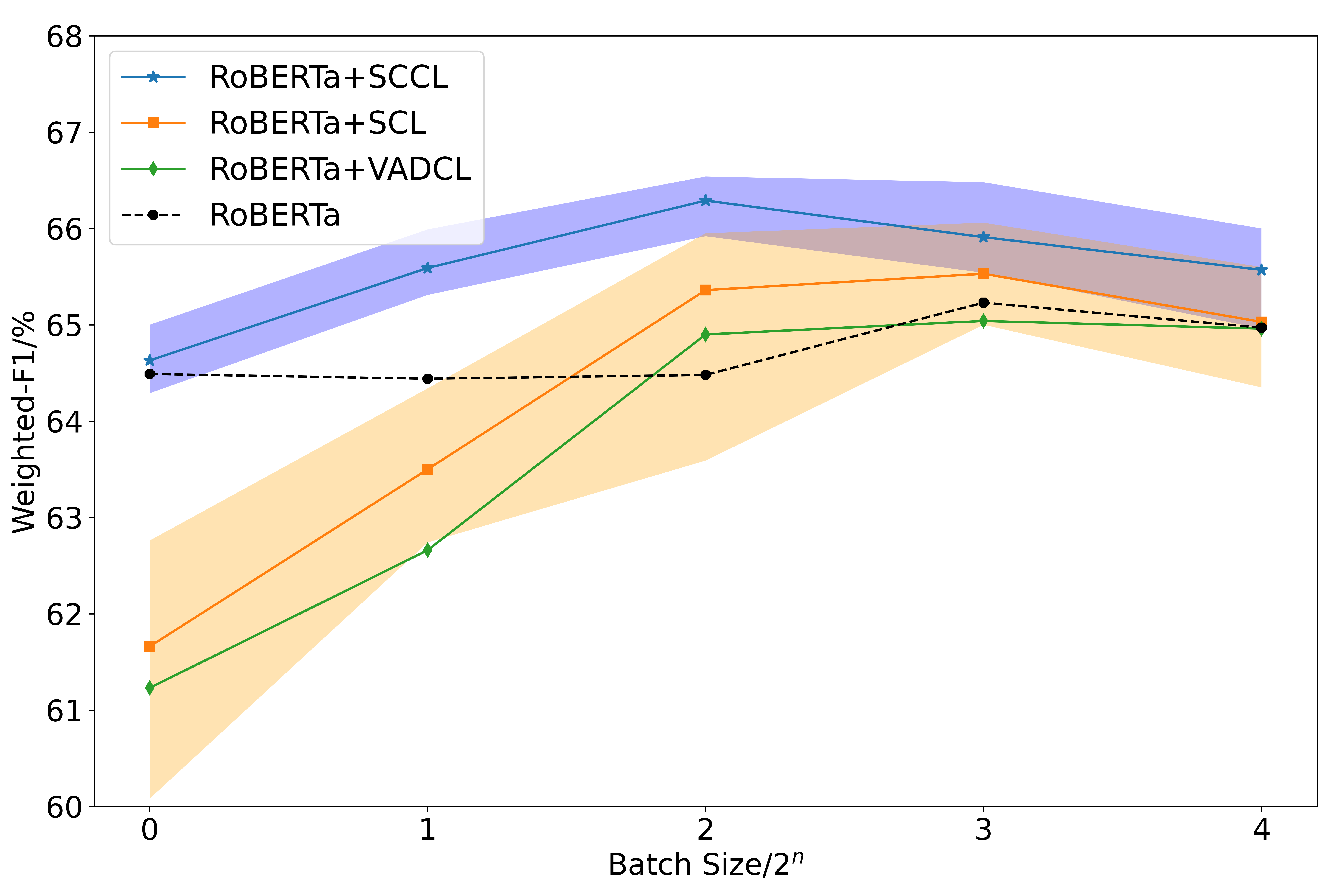}
\caption{Change of F1 scores with different batch sizes on IEMOCAP, using RoBERTa-Base as the encoder.}
\label{fig:line}
\end{figure}

\begin{figure*}[h]
\centering
\includegraphics[width=16cm,height=16cm]{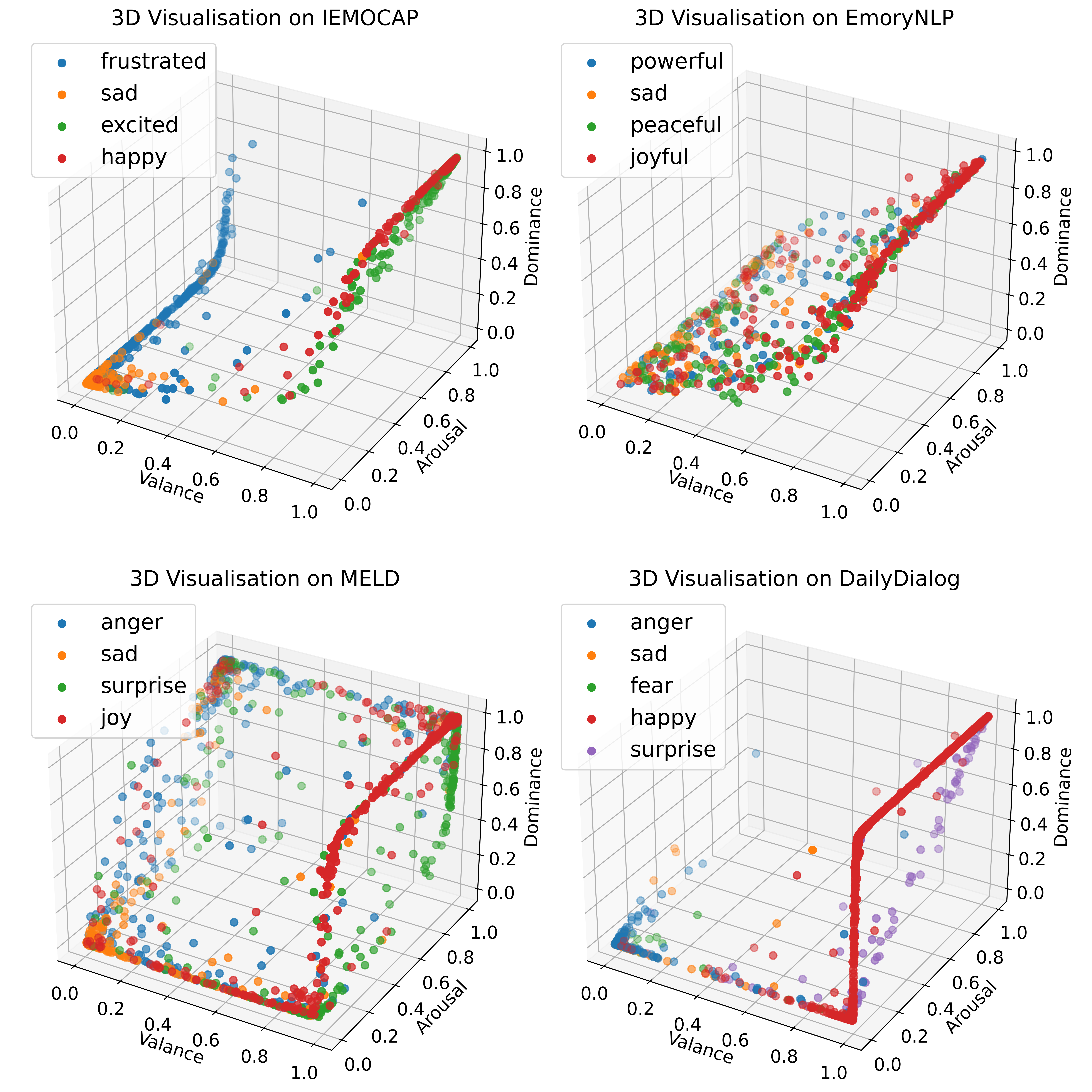}
\caption{Key elements of the VAD visualisation results on all test sets. We only present the samples of representative emotions to provide a more intuitive view.}
\label{fig:3d}
\end{figure*}

\subsection{Batch Size Stability}
\label{batch}
With the change in batch size, we compare the training stability of SCCL, VADCL and SCL in Weighted-F1 scores on IEMOCAP. The results are shown in Figure \ref{fig:line}. Due to the limitation in computational resources, we utilise RoBERTa-Base as encoder and range the batch size from $2^0=1$ to $2^4=16$.

According to the results, RoBERTa achieves the most stable outcomes as the batch size changes, with a Standard Derivation (SD) of 0.32$\%$. SCCL obtains 0.82$\%$ better results than RoBERTa on average, and performs stable as the batch size change, with a SD of 0.66$\%$. This result shows that emotion prototypes obtained from NRC-VAD provide a fixed clustering direction for samples of each emotion. Therefore, the model does not need a large amount of observations at each training step for a stable convergence. 

For SCL, while the results remain competitive and stable with large batch sizes, the performance drops fast below the RoBERTa baseline as the batch size decreases, leading to a high SD of 1.40$\%$. In the extreme case where the batch size drops to 1, SCL fails to converge and brings noise to the training process, resulting in a severe 2.83$\%$ drop compared to RoBERTa. We also provide the variation scale at each batch size for SCCL and SCL. The results show that SCCL has relatively low variances compared to SCL, especially with small batch sizes. This result shows the benefit of NRC-VAD emotion prototypes and the low-dimensional contrastive space, which relieves the curse of dimensionality problem. 

VADCL suffers from the similar problems as SCL, with the highest SD of 1.67$\%$. In addition, VADCL performs worse than SCL with small batch sizes. When observing only a few samples at each training step, the model fails to extract effective features in the three-dimensional space without emotion prototypes as the guidance.

\subsection{Visualisation in VAD Space}
With the three-dimensional VAD space, we are able to directly visualise the predictions instead of utilising dimension reduction techniques. Each VAD prediction also reflects the model's corresponding emotion reasoning process from the Valence-Arousal-Dominance perspective, which benefits interpretability. We present the key elements of the visualisation results on all four test sets in Figure \ref{fig:3d}.

For IEMOCAP, we select and present semantically similar emotions (e.g., \emph{excited} and \emph{happy}) to gain clearer insights to the effect of SCCL, demonstarting their relationships to each other. For dissimilar emotions such as \emph{happy} and \emph{sad}, Valence alone separates them well enough. In addition, similar emotions are also well distinguished in VAD space by Arousal and Dominance, which corresponds with our early hypothesis. For example, \emph{frustrated} and \emph{sad} significantly vary in terms of Arousal, and \emph{happy} and \emph{excited} are jointly divided by Arousal and Dominance.

In section \ref{overall}, we speculate that SCCL provides less improvement to EmoryNLP due to the fuzzy emotions where the VAD prototypes vary in different situations. In the visualisation on EmoryNLP, we present the two fuzzy emotions \emph{powerful}, \emph{peaceful} and two relatively invariant emotions \emph{joyful}, \emph{sad} to provide an intuitive comparison. According to the results, the model makes accurate and well-clustered VAD predictions for samples of \emph{joyful} and \emph{sad}, while the predictions of \emph{peaceful} and \emph{powerful} spread across the VAD space and fail to cluster.

The visualisation results of MELD and DailyDialog shows similar well-separated samples of emotions, such as \emph{joy}/\emph{happy} and \emph{surprise}. However, the predictions of several emotions are inaccurate and not well clustered (e.g., \emph{anger} and \emph{sad} in MELD, \emph{fear} in DailyDialog). We notice that the label distribution of both MELD and DailyDialog is highly imbalanced. Training samples of \emph{sad} cover merely $6.8\%$ in MELD. In DailyDialog, over 60$\%$ of utterances are labelled with \emph{neutral} or \emph{happy}, while the ratio of \emph{fear} and \emph{sad} are both below 5$\%$. Therefore, another direction of future work is to handle the lack of training samples caused by label imbalance for SCCL. In addition, emotions such as \emph{anger} and \emph{sad} are often expressed implicitly, which is closely dependent on the context. Therefore, the lack of contextual information in MELD and DailyDialog brings more challenges to the prediction of these emotions. Overall, the above visualisation results correspond with other experimental outcomes.

\section{Conclusion}
In this paper, we propose a low-dimensional supervised cluster-level contrastive learning model for emotion recognition in conversations. We reduce the high-dimensional supervised contrastive learning space to a three-dimensional space Valence-Arousal-Dominance, and incorporate VAD prototypes from the emotion lexicon NRC-VAD by proposing the novel SCCL method. In addition, we infuse linguistic knowledge and factual knowledge into the context-aware utterance encoder by utilising the pre-trained knowledge adapters.

Experimental results show that our method achieves new state-of-the-art results on three datasets IEMOCAP, MELD, and DailyDialog. Ablation study proves the effectiveness of each proposed module, and further analysis indicates that VAD space is an appropriate and interpretable space for SCCL. Emotion prototypes from NRC-VAD provide useful quantitative information to guide SCCL, which improves model performance and stabilises the training process. The knowledge infused by pre-trained knowledge adapters also enhances the performance of the utterance encoder and SCCL. In the future, we will leverage more fine-grained supervision signals to handle fuzzy emotions, and develop efficient methods to alleviate label imbalance and lack of context problems for SCCL.

\ifCLASSOPTIONcompsoc
  \section*{Acknowledgments}
\else
  \section*{Acknowledgment}
\fi

This research is supported in part by funds from BBSRC, Japan Partnering Award, BB/P025684/1 and MRC, United Kingdom MR/R022461/1. This work is also supported by the Artificial Intelligence Research Center (AIRC), Japan and the University of Manchester President’s Doctoral Scholar award.

\ifCLASSOPTIONcaptionsoff
  \newpage
\fi



%
\bibliographystyle{IEEEtran}
\bibliography{egbib}
%




\end{document}